\documentclass[10pt, a4paper]{article}
\usepackage{lrec2022} 
\usepackage{multibib}
\newcites{languageresource}{Language Resources}
\usepackage{graphicx}
\usepackage{tabularx}
\usepackage{soul}
\usepackage{titlesec}
\titleformat{\section}{\normalfont\large\bfseries\center}{\thesection.}{1em}{}
\titleformat{\subsection}{\normalfont\SmallTitleFont\bfseries\raggedright}{\thesubsection.}{1em}{}
\titleformat{\subsubsection}{\normalfont\normalsize\bfseries\raggedright}{\thesubsubsection.}{1em}{}
\renewcommand\thesection{\arabic{section}}
\renewcommand\thesubsection{\thesection.\arabic{subsection}}
\renewcommand\thesubsubsection{\thesubsection.\arabic{subsubsection}}

\usepackage{epstopdf}
\usepackage[utf8]{inputenc}

\usepackage{hyperref}
\usepackage{xstring}

\usepackage{color}

\usepackage{booktabs}
\usepackage{CJKutf8}
\usepackage{tabularx}
\usepackage{ltablex}
\usepackage{graphicx}
\usepackage{multirow}
\usepackage{microtype}
\usepackage{multicol}
\usepackage{footmisc}
\usepackage{ctable}
\usepackage{amsmath}
\usepackage{newtxtext,newtxmath}
\usepackage[labelfont=bf]{caption}
\usepackage{subcaption}
\usepackage{mathrsfs}
\usepackage{adjustbox}
\usepackage{bbding}
\usepackage{pifont}
\usepackage{soul}

\title{CI-AVSR: A Cantonese Audio-Visual Speech Dataset \\for In-car Command Recognition}

\name{Wenliang Dai$^\star$\thanks{$^\star$ These authors contributed equally.}, Samuel Cahyawijaya$^\star$, Tiezheng Yu$^\star$, Elham J. Barezi$^\star$, \\ \large{\textbf{Peng Xu$^\dagger$\thanks{$^\dagger$ The work was done when the author was studying in The Hong Kong University of Science and Technology.}, Cheuk Tung Shadow Yiu, Rita Frieske, Holy Lovenia,}} \\ \large{\textbf{Genta Indra Winata$^\dagger$, Qifeng Chen, Xiaojuan Ma, Bertram E. Shi, Pascale Fung}}}
\address{The Hong Kong University of Science and Technology \\
\texttt{\{wdaiai,scahyawijaya,tyuah,eebarezi\}@connect.ust.hk}
}

\abstract{
With the rise of deep learning and intelligent vehicles, the smart assistant has become an essential in-car component to facilitate driving and provide extra functionalities. In-car smart assistants should be able to process general as well as car-related commands and perform corresponding actions, which eases driving and improves safety. However, there is a data scarcity issue for low resource languages, hindering the development of research and applications. In this paper, we introduce a new dataset, \textbf{C}antonese \textbf{I}n-car \textbf{A}udio-\textbf{V}isual \textbf{S}peech \textbf{R}ecognition (CI-AVSR), for in-car command recognition in the Cantonese language with both video and audio data. It consists of 4,984 samples (8.3 hours) of 200 in-car commands recorded by 30 native Cantonese speakers. Furthermore, we augment our dataset using common in-car background noises to simulate real environments, producing a dataset 10 times larger than the collected one. We provide detailed statistics of both the clean and the augmented versions of our dataset. Moreover, we implement two multimodal baselines to demonstrate the validity of CI-AVSR. Experiment results show that leveraging the visual signal improves the overall performance of the model. Although our best model can achieve a considerable quality on the clean test set, the speech recognition quality on the noisy data is still inferior and remains as an extremely challenging task for real in-car speech recognition systems. The dataset and code will be released at \url{https://github.com/HLTCHKUST/CI-AVSR}.
 \\ \newline  }
\begin{document}

\maketitleabstract

\section{Introduction}

The research of intelligent transportation systems covers a broad spectrum, including monitoring the driver’s inattention \cite{haghani2021structural}, traffic monitoring for both surveillance and traffic management \cite{buch2011review}, and video-based lane tracking for smart driving assistance \cite{tang2021review}. More recently, the area of self-driving (or autonomous driving) has been developed rapidly with a lot of attention from both academia~\cite{van2018autonomous,zhou2019hidden} and the industry.

Autonomous driving aims to notably improve driving convenience and safety by automating some of the driving tasks, helping drivers to focus more on driving itself. The fast prosperous growth of artificial intelligence (AI) and deep neural networks has led to the growth of autonomous driving. 
However, there is still a long journey to meet the highest level of automation required for autonomous driving systems to deal with various road environments and noise conditions for various driver languages and abilities \cite{zhou2019hidden}. 

Voice controllable systems (VCS), which lean on Automatic Speech Recognition (ASR) methods, have gained remarkable progress with the advances in deep learning and big data. 
It allows drivers to use simple voice commands to handle complex operations, which is a paramount future demand of advanced driving vehicles \cite{zhou2019hidden}. 
However, one of the largest issues of building such an ASR system, especially for low-resource languages, is the lack of data~\cite{winata2020lrt,winata2020mtl,lovenia2021ascend}, which is crucial for achieving commendable speech recognition quality for a VCS system. On the other hand, the audio-visual speech recognition (AVSR) task is proposed to leverage visual data of the speakers to support ASR~\cite{afouras2018deep,Xia2020AudiovisualSR}. It is shown that visual information is beneficial to the ASR, especially when the audio itself is noisy or even unavailable.

To further push the boundary of this research area and mitigate the aforementioned problems, in this paper, we collect a multimodal dataset in the Cantonese language, called Cantonese In-car Audio-Visual Speech Recognition (CI-AVSR), which includes both visual and acoustic signals of the drivers reading various in-car commands. Specifically, it contains 200 unique commands recorded by 30 native Cantonese speakers, forming a dataset of 4,984 samples. Furthermore, we augment it with 10 common in-car background noises to simulate real environments, which increase both the scale and applicability of the dataset. This dataset can benefit the Cantonese and also multilingual AVSR task.

The visual information provided in the CI-AVSR dataset can further enhance the quality of the ASR task. Various research works using audio-visual information~\cite{zadeh2019fmt,dai2020modality,dai2021multimodal} have shown the advantage of utilizing multiple modalities over a single modality. More specifically, recent works in audio-visual speech recognition \cite{zhou2019hidden,259725,alam2020survey,zhu2021deep} have shown that audio-visual information can produce more precise and reliable speech recognition quality compared to only using audio data and can further avoid hidden voice commands that can stealthily control such intelligent systems~\cite{197215}. Moreover, since the in-car voice can be quite noisy, using the visual information helps to enhance the audio and further improve the overall ASR performance.


Overall, our contribution can be summarized as three-fold: 1) we collect and will release the first AVSR dataset for in-car command recognition in the Cantonese language, which could benefit future research and applications; 2) we augment the data with 10 common background noises to simulate real in-car environments and improve the scale; and 3) we provide details statistics of the dataset and train two baseline models to validate its effectiveness.

\section{Related Work}

\begin{CJK*}{UTF8}{bsmi}
\begin{table*}[t!]
\resizebox{\textwidth}{!}{ 
\begin{tabular}{l|l|l|l}
\toprule
Command Category & Command Patterns & Named Entities & Complete Commands \\ \midrule\midrule
Navigation &
  \begin{tabular}[c]{@{}l@{}}1. 導航唔該車我去{[}LOCATION{]}。\\(Please navigate me to [LOCATION], thanks.)\\ 2.  帶我去{[}LOCATION{]}呀，導航。\\(Take me to [LOCATION], navigation.)\\ 3. {[}LOCATION{]}，行邊條路最快去到？\\([LOCATION], what is the fastest path to go there?)\\ 4. 邊條路去{[}LOCATION{]}最近？\\(What is the shortest path to [LOCATION]?)\\ ...\\ 15. {[}LOCATION{]}可唔可以去到？\\(Could you drive me to [LOCATION]?)\end{tabular} &
  \begin{tabular}[c]{@{}l@{}}1. 香港科技大學\\(HKUST)\\ 2. 香港藝術館\\(HK Art Museum)\\ 3. 尖沙咀\\(Tsim Sha Tsui)\\ 4. 海洋公園\\(Ocean Park)\\ ...\\ 15. 維多利亞港\\(Victoria Harbour)\end{tabular} &
  \begin{tabular}[c]{@{}l@{}}1. 導航唔該車我去香港科技大學。\\(Please navigate me to HKUST, thanks.)\\ 2. 導航唔該車我去香港藝術館。\\(Please navigate me to the HK art museum, thanks.)\\ 3. 帶我去尖沙咀呀，導航。\\(Take me to Tsim Sha Tsui, navigation.)\\ 4. 海洋公園，行邊條路最快去到？\\(The ocean park, what is the fastest path to go there?)\\ ...\\ 225. 維多利亞港可唔可以去到？\\(Could you drive me to the Victoria harbour?)\end{tabular} \\ \midrule
Music Playing &
  \begin{tabular}[c]{@{}l@{}}1. 播放{[}SINGER{]}的{[}SONG{]}。\\(Play [SINGER]'s [SONG].)\\ 2. 我想聽{[}SONG{]}。\\(I'd like to listen [SONG].)\\ 3. 我想聽{[}SINGER{]}唱嘅\includegraphics[width = 0.023\textwidth, height = 0.018\textheight]{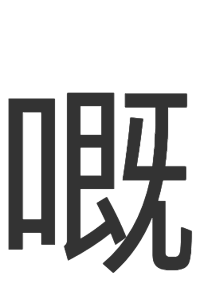}歌。\\(I want to hear a song by [SINGER].)\\ 4. {[}SINGER{]}啲歌唔錯，播嚟\includegraphics[width = 0.023\textwidth, height = 0.021\textheight]{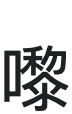}聽下。\\(The songs by [SINGER] are good, play some please.)\\ ...\\ 12. 嚟\includegraphics[width = 0.023\textwidth, height = 0.021\textheight]{images/li.png}首{[}SINGER{]}的{[}SONG{]}。\\(Play the [SONG] by [SINGER].)\end{tabular} &
  \begin{tabular}[c]{@{}l@{}}(SONG/SINGER)\\ 1. 我/張國榮 \\ 2. 海闊天空/Beyond\\ 3. 富士山下/陳奕迅\\ 4. 亂世巨星/陳小春\\ ...\\ 20. 紅日/李克勤\end{tabular} &
  \begin{tabular}[c]{@{}l@{}}1. 播放張國榮的我。\\ 2. 播放Beyond的海闊天空。\\ 3. 我想聽陳奕迅唱嘅\includegraphics[width = 0.023\textwidth, height = 0.018\textheight]{images/kai.png}歌。\\ 4. 嚟\includegraphics[width = 0.023\textwidth, height = 0.021\textheight]{images/li.png}首陳小春的亂世巨星。\\ ...\\ 240. 李克勤啲歌唔錯，播嚟\includegraphics[width = 0.023\textwidth, height = 0.021\textheight]{images/li.png}聽下。\end{tabular} \\ \midrule
Weather Inquiry &
  \begin{tabular}[c]{@{}l@{}} 1. {[}TIME{]}天氣如何？\\(What's the forecast for [TIME]?)\\ 2. 想睇下{[}TIME{]}嘅\includegraphics[width = 0.023\textwidth, height = 0.018\textheight]{images/kai.png}天氣點。\\(I'd like to know the weather on [TIME].)\\ 3. 幫我查下{[}TIME{]}嘅\includegraphics[width = 0.023\textwidth, height = 0.018\textheight]{images/kai.png}天氣。\\(Please help me to check the weather [TIME].)\\ 4. 唔該講下{[}TIME{]}天氣點啊？\\(What's the weather [TIME]? Thanks.)\\ ...\\ 7. {[}TIME{]}天氣好啲，定係{[}TIME{]}天氣好啲？\\(The weather [TIME] seems to be good, is it?)\end{tabular} &
  \begin{tabular}[c]{@{}l@{}}1. 明天\\(Tomorrow)\\ 2. 今天晚上\\(Tonight)\\ 3. 星期三\\(Wednesday)\\ 4. 禮拜天\\(Sunday)\\ ...\\ 10. 週六\\(Saturday)\end{tabular} &
  \begin{tabular}[c]{@{}l@{}}1. 明天天氣如何？\\(What's the forecast for tomorrow?)\\ 2. 今天晚上天氣如何？\\(What's the forecast for tonight?)\\ 3. 唔該講下星期三天氣點啊？\\(What's the weather on Wednesday? Thanks.)\\ 4. 幫我查下禮拜天嘅\includegraphics[width = 0.023\textwidth, height = 0.018\textheight]{images/kai.png}天氣。\\(Please help me to check the weather on Sunday.)\\ ...\\ 70. 週六天氣好啲，定係週六天氣好啲？\\(The weather on Saturday seems to be good, is it?)\end{tabular} \\ \bottomrule
\end{tabular}
}
\caption{Examples of command patterns, named entities, and the combination of them to form complete commands for the three categories, including \textit{navigation}, \textit{music playing}, and \textit{weather inquiry}. English translations are provided in the parentheses, except the singers and songs that cannot be translated directly.}
\label{tab:commands_template}
\end{table*}
\end{CJK*}

\subsection{AVSR and In-car Datasets}
In the past few years, many multimodal datasets have been released to facilitate the study of in-car speech recognition, driver behavior, emotion recognition, etc. \newcite{jenness2008use} conduct an empirical research study about the use of the voice control system (VCS) by drivers as well as potential measures that could be used for evaluating possible distraction from using these systems while driving a car. To this end, they collect 30 minutes of data on car-driving on two US roads to explore the performance of vehicle voice control interface. The data cover various situations, including the highway, heavy arterials, residential, and commercial streets.



On the other hand, \newcite{ivanecky2012car} introduce a speech dataset in German for addressing the ASR task for physically disabled drivers. This dataset is recorded by 10 speakers and each of them is asked to record $2 \times 30$ commands. Furthermore, to include environmental noises, the speakers conduct the data recording at a distance of 20 to 30 centimeters from the microphone.

While in the AVSR task, \newcite{afouras2018deep} have proposed a large-scale dataset named LRS2-BBC, which is collected from the British Broadcasting Corporation (BBC). The dataset contains thousands of hours of spoken sentences from various BBC programs that contain talking faces together with transcripts. Moreover, \newcite{afouras2018lrs3} further introduce a multimodal dataset for visual and audio-visual speech recognition. It includes face tracks from over 400 hours of TED and TEDx videos, along with the
corresponding subtitles and word alignment boundaries. 

More recently, \cite{stappen2021multimodal} propose the MuSe-Sent challenge that aims to predict five sentiment classes for each emotion dimension (arousal or valence) on a segment level, based on audio-visual recordings from the MuSe-CaR dataset about car review videos. The dataset is crawled from YouTube which contains 291 videos. However, to the best of our knowledge, none of the previous work focuses on in-car command AVSR for low-resource languages, such as Cantonese, which hinders the research and applications of AVSR in those languages.


\subsection{AVSR Models}
Various architectures have been proposed to tackle the AVSR task. \newcite{petridis2018end} use bidirectional GRUs to encode the visual and acoustic data separately and adopt a late fusion for two modalities. It shows that even with a simple model, visual information can improve the performance. \newcite{afouras2018deep} propose to leverage the Transformer self-attention and compare the effectiveness of two objectives: the Connectionist Temporal Classification (CTC) loss and the sequence-to-sequence language modeling loss. They show that visual information can further reduce the word error rate up to 1.2\% in a clean environment, and 23.2\% in a noisy environment on a very large-scale dataset. It indicates that the visual information of lip movement can benefit a lot when the audio is noisy. 

In recent years, the use of attention-based models has been increasing for solving the AVSR task. \newcite{xu2020discriminative} propose a double visual awareness
multi-modality speech recognition (AE-MSR) network to use the visual information with two steps, first to enhance the noisy audio data, and then as a second modality using an Element-wise-Attention Gated Recurrent Unit (EleAtt-GRU), which is potentially more effective than Transformer for long sequences. More recently, \newcite{ma2021end} introduce a novel architecture by combining convolutional layers with Transformer, achieving state-of-the-art performance on common benchmarks. In this paper, we build two multimodal baseline models to demonstrate the validity of our collected dataset.

\section{Dataset Collection and Statistics}
In this section, we first describe the data collection pipeline for CI-AVSR, which includes the command template creation, the data recording interface, the recording equipment and format, and the data recording strategy. Then, we provide detailed statistics of the CI-AVSR dataset. Finally, we introduce a data augmentation method to increase the data scale and simulate in-car environments.

\subsection{The Template of In-car Commands}
Before the data recording phase, we first prepare a template of commands for the speakers to read, in which the commands should be diverse and cover common scenarios. 
We classify template of commands into four general categories: 1) \textit{navigation}; 2) \textit{music playing}; 3) \textit{weather inquiry}; and 4) \textit{others}. This categorization strategy allows us to cover all the common scenarios of the in-car voice commands while maintaining the diversity of the commands and further simplifying the commands creation process.

For the first three categories, we construct the template of commands with three steps. Firstly, for each category, we construct multiple command patterns without specifying named entities, but instead, we use placeholder tags. For the  \textit{navigation}, we use a placeholder tag \texttt{[LOCATION]}, for the \textit{music playing}, we use placeholder tags \texttt{[SINGER]} and \texttt{[SONG]}, and for the \textit{weather inquiry}, we use a placeholder tag \texttt{[TIME]}. We then replace the placeholder tag with a suitably named entity to fill in the command patterns. Finally, for each command pattern, we iteratively infill all the placeholder tags with name entities to form complete commands. Examples of this procedure are shown in Table~\ref{tab:commands_template}. 

In addition to the first three categories, there are also other frequently used in-car commands. However, these commands cannot be created by following the aforementioned procedure. We categorize these commands as \textit{others}, including commands to ask the system to turn on or off some functionality in the car (e.g., air-conditioner, radio, light, etc.), ask the system to broadcast or send messages, and inquiry the status of the car (e.g., how much electricity or gas remains). 

To ensure the created commands conform to the spoken Cantonese language, we hire two human experts who are native Cantonese speakers. First of all, each expert is asked to design command candidates for all categories (command patterns and named entities for the first three, and complete commands for the last). Then, we swap their designed results and ask them to conduct a grammar and correctness check for each other. Finally, we assemble all the commands and ask these two experts to filter out command patterns with high similarities to increase the diversity. Statistics of these preliminary commands are shown in Table~\ref{tab:commands_stats}. 

\begin{table}[t]
\centering
\resizebox{\linewidth}{!}{ 
\begin{tabular}{lccc}
\toprule
Command Category & \multicolumn{1}{l}{\#Patterns} & \multicolumn{1}{l}{\#Entities} & \multicolumn{1}{l}{\#Commands} \\ \midrule\midrule
Navigation      & 15 & 15 & 225 \\
Music Playing   & 12 & 20 & 240 \\
Weather Inquiry & 7  & 10 & 70  \\ 
Others & -  & - & 40  \\ \bottomrule
\end{tabular}
}
\caption{Statistics of commands among different categories before the sampling.}
\label{tab:commands_stats}
\end{table}

\begin{figure}[t!]
     \centering
    \includegraphics[width=1.0\linewidth]{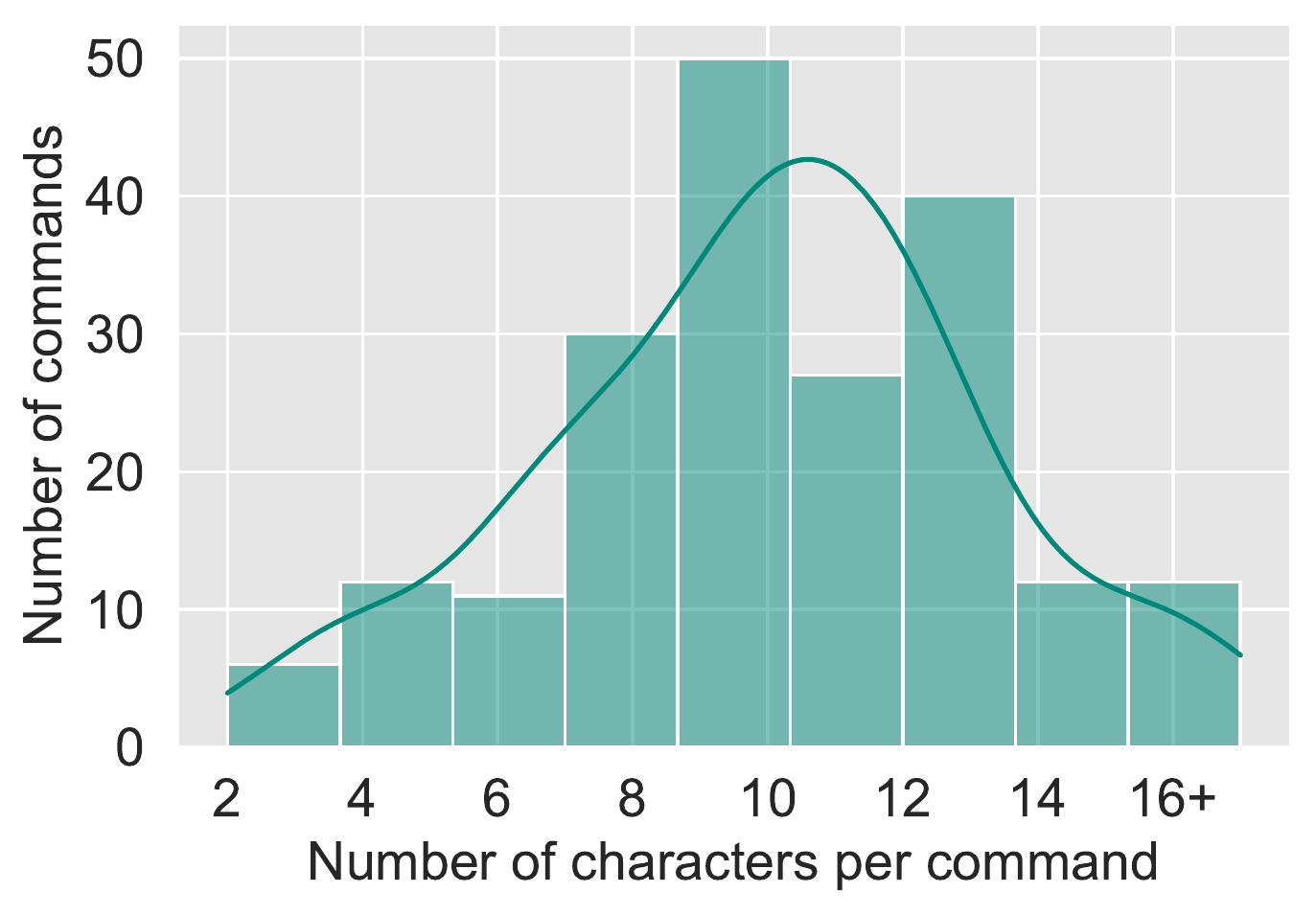}
    \caption{Distribution of the number of characters in all commands. For commands that contains English words, each English word is counted as one character. The last column on the right represents commands that have more than 16 characters.}
    \label{fig:num_chars}
\end{figure}

One drawback of creating commands in this way is that many commands share the same pattern with only a small difference in the named entities, which reduces the overall diversity of the commands. To mitigate this problem, we uniformly sample $\sim$30\% commands from the first three categories while keeping all the commands from the \textit{others} category. Finally, after sampling, we collect  200 commands in total, in which 160 are from the first three categories and 40 are from the \textit{others} category\footnote{The full command list will be released along with our dataset in \url{https://github.com/HLTCHKUST/CI-AVSR}.}. We show the distribution of command length in Figure~\ref{fig:num_chars}. 

\subsection{Data Recording} \label{sec:data_collection}

\begin{figure}[t]
    \centering
    \includegraphics[width=\linewidth]{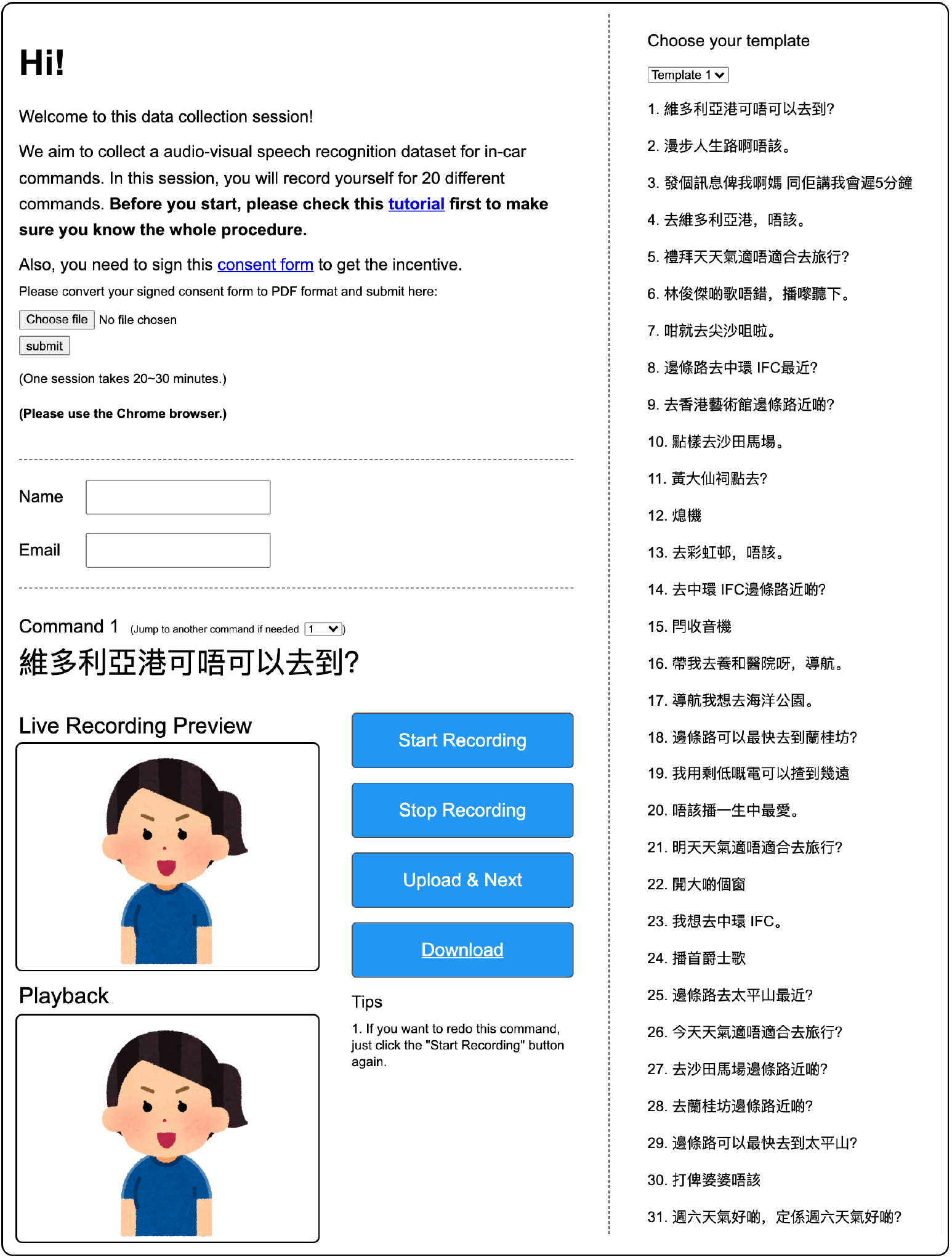}
    \caption{The web interface for conducting data recording. It helps us to distribute works and reduce the time cost. Details are explained in Section~\ref{sec:data_collection}.}
    \label{fig:interface}
\end{figure}

\paragraph{Data Recording Interface.} To reduce the difficulty of data recording, we make a dedicated website so that speakers can conduct the recording on their own computer (equipped with a camera and a microphone) at their own place\footnote{This is especially important during the COVID-19 pandemic as people usually work from home and there is a government regulation on the number of people for gatherings.}. The interface of the website is shown in Figure~\ref{fig:interface}. In the top left part, we show a basic introduction to the speakers as well as a tutorial to teach them how to use this website and how to record. The control panel is on the bottom left, where there is the current recording command shown, the buttons to perform actions (start/stop the recording and save/download the recorded data), a live preview window to show the real-time camera view, and a playback window to play the last recorded video. Once the speakers are satisfied with the recording of the current command, they can proceed to the next one until all the commands are recorded. On the right-hand side, we show an overview of all the commands that the speakers need to fulfill, which also shows the current progress.

\paragraph{Data Recording Equipment and Format.} Although the previously mentioned setting is convenient to distribute data collection jobs and reduce the cost, there is a problem that the speakers may have different types of cameras and microphones. To mitigate this problem, we ask the speakers to use a camera with at least 720p (1280 $\times$ 720 px) resolution, which gives us sufficient room for further processing. After the recording, we crop all videos to $640 \times 480$ with the person at the center area. For the recorded audios, we process them to the same format with the 16kHz frame rate, mono channel, and encoded as 16-bit pulse-code modulation (PCM), producing a total bit rate of 256 kbps.

\begin{figure}[t!]
    \centering
    \includegraphics[width=1.0\linewidth]{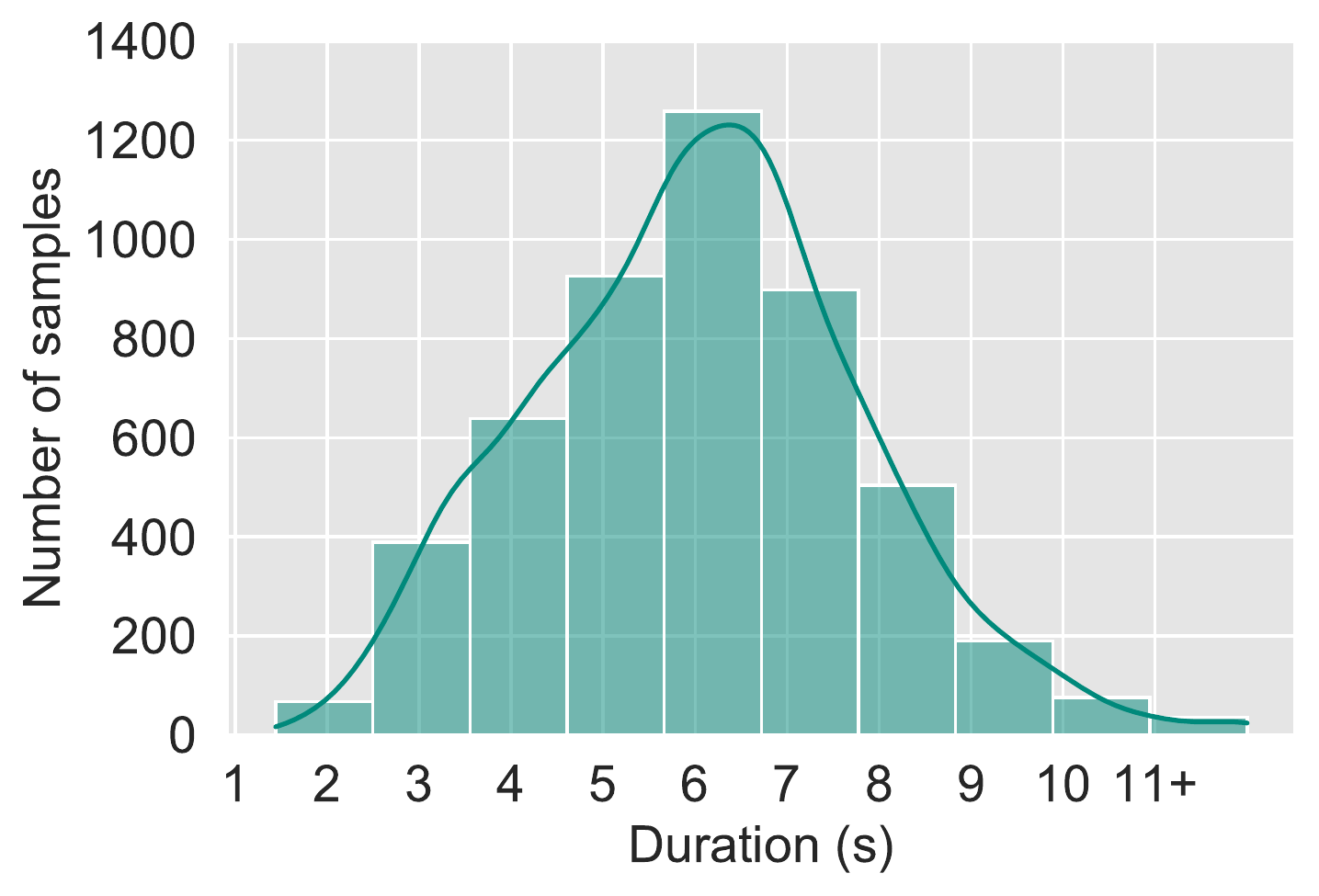}
    \caption{Duration distribution of the recorded data. The last column on the right represents samples that are longer than 11 seconds.}
    \label{fig:duration}
\end{figure}

\paragraph{Data Recording Strategy.} We divide the whole data recording process into two stages. In the first stage, we perform a preliminary data collection session to test the robustness of our system and the correctness of the created commands. To this end, we hire 10 native Cantonese speakers (5 males and 5 females) and ask each of them to record 100 commands. Based on their feedback, we improve the user experience on the website, provide more detailed instructions in the tutorial, and add a more potent data auto-saving strategy of the website to avoid data lost. In the second stage, we expand the data collection scale by hiring 20 native Cantonese speakers (10 males and 10 females) and asking each of them to record the full list of 200 commands. In the total of two stages, there are 5,000 samples collected. We perform a manual check on each of the data samples and filter out 16 samples with low quality. Therefore, we have 4,984 data samples for the final dataset with a total duration of 30,049 seconds. The distribution of sample duration is shown in Figure~\ref{fig:duration}. Finally, as illustrated in Table~\ref{tab:data_split}, we split the data into training, validation, and test sets by speakers while maintaining a balanced gender distribution on each split.


\begin{table}[t]
\centering
\resizebox{0.9\linewidth}{!}{
\begin{tabular}{l|ccc}
\toprule
Split & \#Male (Dur.) & \#Female (Dur.) & Total Dur. \\ \midrule \midrule
Train & 10 (10,803s)   & 10 (11,813s)     & 22,616s     \\
Valid & 2 (1,849s)     & 2 (1,829s)       & 3,678s      \\
Test  & 3 (1,902s)     & 3 (1,843s)       & 3,745s      \\ \bottomrule
\end{tabular}
}
\caption{Statistics of the train/valid/test splits of our recorded dataset. We split data by the speaker id, i.e. speakers in a set will not appear in another. Here, Dur. denotes duration and s represents seconds.}
\label{tab:data_split}
\end{table}

\subsection{Data Augmentation}

To simulate in-car environments and increase the data scale, we augment each sample from the collected dataset by combining it with 10 different in-car background noises\footnote{The sounds are downloaded from \url{https://freesound.org/}.} that are commonly heard in the daily usage of cars, including \textit{alarm}, \textit{horn}, \textit{background music}, \textit{ignition},  \textit{hail}, \textit{rain}, \textit{windscreen wiper}, \textit{road ambience}, \textit{door opens and closes}, and \textit{people talking}. For each type of noise, we use five variants to increase the diversity and uniformly sample one when applying to the data. The volume of the noise is adjusted by human experts so that the original commands are still recognizable and all the sounds are on the same level of loudness. Therefore, the resulting augmented dataset is 10 times as large as the original clean one and more in line with actual in-car scenarios, which could potentially benefit the generalization and applicability of the trained model.

\section{Experiments}
\label{sec:experiments}


In this section, we aim to demonstrate the validity of our collected multimodal Cantonese data and the effectiveness of using the visual signals to help speech recognition. 

\subsection{Baseline models}

\paragraph{Conformer-based multimodal ASR model.} Following \newcite{Ma2021EndToEndAS}, for the audio input, we use a modified ResNet-18~\cite{He2016DeepRL} with all 2D-convolution layers changed to 1D, and then a Convolution-augmented Transformer (Conformer~\cite{Gulati2020ConformerCT}) encoder to further process the representations. For the visual input, it is also first processed by a ResNet-18 with the first layer replaced by a 3D convolutional layer, and then another Conformer encoder to get final visual representations. The audio and visual representations are fused by a Multi-Layer-Perceptron (MLP), which consists of two linear layers with batch normalization and ReLU activation in between. Finally, we apply a GPT-based~\cite{radford2019language} four-layer decoder to decode the audio-video information to text.


\paragraph{Wav2Vec 2.0-based multimodal ASR model.} The pre-trained Wav2Vec 2.0 model~\cite{Baevski2020wav2vec2A} gives strong speech representations and shows excellent performance on downstream tasks. We further extend it with a video encoder to allow audio-visual speech recognition. We adopt a ResNet-18 model pre-trained the ImageNet~\cite{Russakovsky2015ImageNetLS} to get the visual representations from each video frame. Moreover, we incorporate another 1D-convolutional layer with a kernel size of 3 to add interaction between frame-level features. To fuse the representations from two modalities, we align the audio framerate with the video framerate and fuse them by summing representations from both modalities. Lastly, we apply a linear transformation to convert the fused representations into predicted tokens.


\subsection{Implementation Details}

\paragraph{Data Preprocessing}
In our experiment, we preprocess the multimodal data to get a standardized input format of the audio and the image data. For the audio data, we keep using 16kHz mono-channel audio data and normalize the value with zero mean and standard deviation of one. For the image data, we extract 25 image frames per second from the video using FFMPEG~\cite{tomar2006converting}. To reduce the computational cost, we extract only the lip part of each image with the face landmark detection module~\cite{kazemi2014facedetect} from the DLIB library~\cite{king09dlib} by using a face landmark detection model \footnote{\url{http://dlib.net/files/shape_predictor_68_face_landmarks.dat.bz2}} that is pre-trained on the iBUG 300-W face dataset~\cite{sagonas2016ibug} \footnote{\url{https://ibug.doc.ic.ac.uk/resources/facial-point-annotations}}.



\paragraph{Training Details.}
During training, we grayscale and downsize the image samples into $32\times32$ pixels. We use the Adam~\cite{Kingma2015AdamAM} optimizer to optimize all the models in all experiments. For the Conformer-based model, we use a learning rate of $5e^{-5}$, a batch size of 3, and minimize the cross-entropy loss from the output of the transformer decoder. Except the ResNet-18 for extracting the visual features, all other modules in the model are trained from scratch. As for Wav2Vec 2.0-based model, we utilize a pre-trained Wav2Vec 2.0 model which has been fine-tuned on the Cantonese CommonVoice dataset~\footnote{\url{https://huggingface.co/ctl/wav2vec2-large-xlsr-cantonese}}. For the Wav2Vec 2.0 model, we fine-tuned the whole model's parameters using a learning rate of $5e^{-5}$ and a batch size of 16 by minimizing the CTC loss~\cite{Graves06connectionisttemporal} of the output with regard to the label.

\paragraph{Evaluation Details.}
We use the character error rate (CER) \cite{Wang_CER} rather than the word error rate (WER) as an evaluation metric because the Cantonese language is character-based. In detail, the CER is calculated by adding the number of substituted, inserted, and deleted characters together and then dividing them by the total number of characters. For the multimodal Conformer model, the output transcripts are generated in an auto-regressive manner with a beam search size of 4 and a length penalty of 1. For the multimodal wav2vec 2.0 model, the output transcription is generated using the CTC decoding method.

\begin{table}[t]
\begin{adjustbox}{width=\linewidth,totalheight={\textheight},keepaspectratio}
\begin{tabular}{l|cccc}
\toprule
\multirow{2}{*}{Model} & \multicolumn{2}{c}{Audio only (CER)}    & \multicolumn{2}{c}{Audio + Video (CER)}  \\
                       & clean & noisy & clean & noisy \\ \midrule
Conformer              & 31.29\%            & 167.63\%     & 26.97\%            & 132.68\%       \\
Wav2Vec2               & 4.06\%            & 12.75\%     & 3.48\%            & 7.19\%       \\ \bottomrule
\end{tabular}
\end{adjustbox}
\caption{The character error rate (CER) on the test set of our collected data. The two models are only trained on the clean training set and tested on both the clean and augmented (noisy) test sets.}
\label{tab:experiment_results}
\end{table}
\subsection{Results and Analysis}
We train the Conformer-based and Wav2Vec-based models only on the clean training set of our data with two settings: 1) audio-only; and 2) multimodal (audio and video). As shown in Table~\ref{tab:experiment_results}, we evaluate their performance on both clean and augmented (noisy) test sets. Compared to the audio-only setting, models trained on multimodal data achieve lower CER by a large margin on both clean and noisy data. Moreover, similar to prior work~\cite{afouras2018deep,xu2020discriminative,ma2021end}, we find that the visual signal contributes more when the audio is noisy.
The rich visual data provide complementary and supplementary information to the audio data for the models to generate transcripts. However, the performance of the Conformer-based models is far from satisfactory. We conjecture that the model is overfitted to the training set since our clean training data is limited, which could be mitigated by using the augmented data to train. 
On the other hand, the pre-trained Wav2Vec-based model has great generalization ability and shows promising results after finetuning. 
Additionally, to ablate the effects of each background noise, in Table~\ref{tab:noise_perf}, we show the test result of each of them using the Wav2Vec-based model. The model is more sensitive to \textit{alarm} and \textit{horn} compared to the others, and the visual information helps more under these situations.

\begin{table}[]
\centering
\resizebox{\linewidth}{!}{
\begin{tabular}{l|cc}
\toprule
\multirow{2}{*}{Type of Noise} & \multicolumn{2}{c}{CER} \\ \cmidrule{2-3} 
                               & Audio   & Audio + Video \\ \midrule
Clean                          & 4.06\%  & 3.48\%        \\ \midrule
0 (background music)           & 11.53\% & 5.42\%        \\
1 (rain)                       & 12.16\% & 6.19\%        \\
2 (hail)                       & 17.03\% & 10.03\%       \\
3 (ignition)                   & 21.73\% & 13.65\%       \\
4 (windscreen wiper)           & 13.57\% & 8.50\%        \\
5 (horn)                       & 16.26\% & 9.06\%        \\
6 (people talking)             & 13.53\% & 7.78\%        \\
7 (road ambience)              & 13.25\% & 7.36\%        \\
8 (alarm)                      & 17.13\% & 8.58\%        \\
9 (car door)                   & 7.19\%  & 3.38\%        \\ \midrule
Avg (0 to 9)                         & 14.34\% & 7.99\%        \\ \bottomrule
\end{tabular}
}
\caption{The character error rate of the Wav2Vec 2.0 model (trained on the clean training set only) on the augmented noisy data. We report its performance on each type of noise and the average of them.}
\label{tab:noise_perf}
\end{table}

\section{Conclusion}
In this paper, we propose a new dataset, Cantonese In-car Audio-Visual Speech Recognition (CI-AVSR), for audio-visual speech recognition of in-car commands. It consists of 200 unique commands with 8.3 hours of recorded data. Furthermore, we augment the dataset with 10 commonly seen background sounds to simulate real scenarios, resulting in more than 80 hours of data. We evaluate the collected data with two baseline models, showing the effectiveness of AVSR. When testing on the augmented data with background noises, we observe a clear performance drop, which we believe would be a challenging and interesting future research direction to tackle.

\section{Acknowledgements}
This work is funded by ITS/353/19FP of the Innovation Technology Commission, The Hong Kong SAR Government, School of Engineering Ph.D. Fellowship Award, the Hong Kong University of Science and Technology, and the Hong Kong Fellowship Scheme by the Hong Kong Research Grants Council (RGC).

\bibliographystyle{lrec2022-bib}
\bibliography{lrec2022}

\begin{thebibliography}{}

\bibitem[\protect\citename{Afouras \bgroup et al.\egroup
  }2018a]{afouras2018deep}
Afouras, T., Chung, J.~S., Senior, A., Vinyals, O., and Zisserman, A.
\newblock (2018a).
\newblock Deep audio-visual speech recognition.
\newblock {\em IEEE transactions on pattern analysis and machine intelligence}.

\bibitem[\protect\citename{Afouras \bgroup et al.\egroup
  }2018b]{afouras2018lrs3}
Afouras, T., Chung, J.~S., and Zisserman, A.
\newblock (2018b).
\newblock Lrs3-ted: a large-scale dataset for visual speech recognition.
\newblock {\em arXiv preprint arXiv:1809.00496}.

\bibitem[\protect\citename{Alam \bgroup et al.\egroup }2020]{alam2020survey}
Alam, M., Samad, M.~D., Vidyaratne, L., Glandon, A., and Iftekharuddin, K.~M.
\newblock (2020).
\newblock Survey on deep neural networks in speech and vision systems.
\newblock {\em Neurocomputing}, 417:302--321.

\bibitem[\protect\citename{Baevski \bgroup et al.\egroup
  }2020]{Baevski2020wav2vec2A}
Baevski, A., Zhou, H., rahman Mohamed, A., and Auli, M.
\newblock (2020).
\newblock wav2vec 2.0: A framework for self-supervised learning of speech
  representations.
\newblock {\em ArXiv}, abs/2006.11477.

\bibitem[\protect\citename{Buch \bgroup et al.\egroup }2011]{buch2011review}
Buch, N., Velastin, S.~A., and Orwell, J.
\newblock (2011).
\newblock A review of computer vision techniques for the analysis of urban
  traffic.
\newblock {\em IEEE Transactions on intelligent transportation systems},
  12(3):920--939.

\bibitem[\protect\citename{Carlini \bgroup et al.\egroup }2016]{197215}
Carlini, N., Mishra, P., Vaidya, T., Zhang, Y., Sherr, M., Shields, C., Wagner,
  D., and Zhou, W.
\newblock (2016).
\newblock Hidden voice commands.
\newblock In {\em 25th USENIX Security Symposium (USENIX Security 16)}, pages
  513--530, Austin, TX, August. USENIX Association.

\bibitem[\protect\citename{Dai \bgroup et al.\egroup }2020]{dai2020modality}
Dai, W., Liu, Z., Yu, T., and Fung, P.
\newblock (2020).
\newblock Modality-transferable emotion embeddings for low-resource multimodal
  emotion recognition.
\newblock In {\em Proceedings of the 1st Conference of the Asia-Pacific Chapter
  of the Association for Computational Linguistics and the 10th International
  Joint Conference on Natural Language Processing}, pages 269--280, Suzhou,
  China, December. Association for Computational Linguistics.

\bibitem[\protect\citename{Dai \bgroup et al.\egroup }2021]{dai2021multimodal}
Dai, W., Cahyawijaya, S., Liu, Z., and Fung, P.
\newblock (2021).
\newblock Multimodal end-to-end sparse model for emotion recognition.
\newblock In {\em Proceedings of the 2021 Conference of the North American
  Chapter of the Association for Computational Linguistics: Human Language
  Technologies}, pages 5305--5316, Online, June. Association for Computational
  Linguistics.

\bibitem[\protect\citename{Graves \bgroup et al.\egroup
  }2006]{Graves06connectionisttemporal}
Graves, A., Fernández, S., and Gomez, F.
\newblock (2006).
\newblock Connectionist temporal classification: Labelling unsegmented sequence
  data with recurrent neural networks.
\newblock In {\em In Proceedings of the International Conference on Machine
  Learning, ICML 2006}, pages 369--376.

\bibitem[\protect\citename{Gulati \bgroup et al.\egroup
  }2020]{Gulati2020ConformerCT}
Gulati, A., Qin, J., Chiu, C.-C., Parmar, N., Zhang, Y., Yu, J., Han, W., Wang,
  S., Zhang, Z., Wu, Y., and Pang, R.
\newblock (2020).
\newblock Conformer: Convolution-augmented transformer for speech recognition.
\newblock {\em ArXiv}, abs/2005.08100.

\bibitem[\protect\citename{Haghani \bgroup et al.\egroup
  }2021]{haghani2021structural}
Haghani, M., Behnood, A., Oviedo-Trespalacios, O., and Bliemer, M.~C.
\newblock (2021).
\newblock Structural anatomy and temporal trends of road accident research:
  Full-scope analyses of the field.
\newblock {\em Journal of safety research}.

\bibitem[\protect\citename{He \bgroup et al.\egroup }2016]{He2016DeepRL}
He, K., Zhang, X., Ren, S., and Sun, J.
\newblock (2016).
\newblock Deep residual learning for image recognition.
\newblock {\em 2016 IEEE Conference on Computer Vision and Pattern Recognition
  (CVPR)}, pages 770--778.

\bibitem[\protect\citename{Ivaneck{\`y} and Mehlhase}2012]{ivanecky2012car}
Ivaneck{\`y}, J. and Mehlhase, S.
\newblock (2012).
\newblock An in-car speech recognition system for disabled drivers.
\newblock In {\em International Conference on Text, Speech and Dialogue}, pages
  505--512. Springer.

\bibitem[\protect\citename{Jenness \bgroup et al.\egroup }2008]{jenness2008use}
Jenness, J.~W., Lerner, N.~D., Mazor, S., Osberg, J.~S., and Tefft, B.~C.
\newblock (2008).
\newblock Use of advanced in-vehicle technology by young and older early
  adopters.
\newblock {\em Survey results on adaptive cruise control systems. Report no.
  DOT HS}, 810:917.

\bibitem[\protect\citename{Kazemi and Sullivan}2014]{kazemi2014facedetect}
Kazemi, V. and Sullivan, J.
\newblock (2014).
\newblock One millisecond face alignment with an ensemble of regression trees.
\newblock In {\em 2014 {IEEE} Conference on Computer Vision and Pattern
  Recognition}. {IEEE}, June.

\bibitem[\protect\citename{King}2009]{king09dlib}
King, D.~E.
\newblock (2009).
\newblock Dlib-ml: A machine learning toolkit.
\newblock {\em Journal of Machine Learning Research}, 10:1755--1758.

\bibitem[\protect\citename{Kingma and Ba}2015]{Kingma2015AdamAM}
Kingma, D.~P. and Ba, J.
\newblock (2015).
\newblock Adam: A method for stochastic optimization.
\newblock {\em CoRR}, abs/1412.6980.

\bibitem[\protect\citename{Lovenia \bgroup et al.\egroup
  }2021]{lovenia2021ascend}
Lovenia, H., Cahyawijaya, S., Winata, G.~I., Xu, P., Yan, X., Liu, Z., Frieske,
  R., Yu, T., Dai, W., Barezi, E.~J., and Fung, P.
\newblock (2021).
\newblock Ascend: A spontaneous chinese-english dataset for code-switching in
  multi-turn conversation.

\bibitem[\protect\citename{Ma \bgroup et al.\egroup }2021a]{ma2021end}
Ma, P., Petridis, S., and Pantic, M.
\newblock (2021a).
\newblock End-to-end audio-visual speech recognition with conformers.
\newblock In {\em ICASSP 2021-2021 IEEE International Conference on Acoustics,
  Speech and Signal Processing (ICASSP)}, pages 7613--7617. IEEE.

\bibitem[\protect\citename{Ma \bgroup et al.\egroup }2021b]{Ma2021EndToEndAS}
Ma, P., Petridis, S., and Pantic, M.
\newblock (2021b).
\newblock End-to-end audio-visual speech recognition with conformers.
\newblock {\em ICASSP 2021 - 2021 IEEE International Conference on Acoustics,
  Speech and Signal Processing (ICASSP)}, pages 7613--7617.

\bibitem[\protect\citename{Petridis \bgroup et al.\egroup
  }2018]{petridis2018end}
Petridis, S., Stafylakis, T., Ma, P., Cai, F., Tzimiropoulos, G., and Pantic,
  M.
\newblock (2018).
\newblock End-to-end audiovisual speech recognition.
\newblock In {\em 2018 IEEE international conference on acoustics, speech and
  signal processing (ICASSP)}, pages 6548--6552. IEEE.

\bibitem[\protect\citename{Radford \bgroup et al.\egroup
  }2019]{radford2019language}
Radford, A., Wu, J., Child, R., Luan, D., Amodei, D., Sutskever, I., et~al.
\newblock (2019).
\newblock Language models are unsupervised multitask learners.
\newblock {\em OpenAI blog}, 1(8):9.

\bibitem[\protect\citename{Russakovsky \bgroup et al.\egroup
  }2015]{Russakovsky2015ImageNetLS}
Russakovsky, O., Deng, J., Su, H., Krause, J., Satheesh, S., Ma, S., Huang, Z.,
  Karpathy, A., Khosla, A., Bernstein, M.~S., Berg, A.~C., and Fei-Fei, L.
\newblock (2015).
\newblock Imagenet large scale visual recognition challenge.
\newblock {\em International Journal of Computer Vision}, 115:211--252.

\bibitem[\protect\citename{Sagonas \bgroup et al.\egroup
  }2016]{sagonas2016ibug}
Sagonas, C., Antonakos, E., Tzimiropoulos, G., Zafeiriou, S., and Pantic, M.
\newblock (2016).
\newblock 300 faces in-the-wild challenge: database and results.
\newblock {\em Image and Vision Computing}, 47:3--18, March.

\bibitem[\protect\citename{Stappen \bgroup et al.\egroup
  }2021]{stappen2021multimodal}
Stappen, L., Baird, A., Schumann, L., and Schuller, B.
\newblock (2021).
\newblock The multimodal sentiment analysis in car reviews (muse-car) dataset:
  Collection, insights and improvements.
\newblock {\em arXiv preprint arXiv:2101.06053}.

\bibitem[\protect\citename{Tang \bgroup et al.\egroup }2021]{tang2021review}
Tang, J., Li, S., and Liu, P.
\newblock (2021).
\newblock A review of lane detection methods based on deep learning.
\newblock {\em Pattern Recognition}, 111:107623.

\bibitem[\protect\citename{Tomar}2006]{tomar2006converting}
Tomar, S.
\newblock (2006).
\newblock Converting video formats with ffmpeg.
\newblock {\em Linux Journal}, 2006(146):10.

\bibitem[\protect\citename{Van~Brummelen \bgroup et al.\egroup
  }2018]{van2018autonomous}
Van~Brummelen, J., O’Brien, M., Gruyer, D., and Najjaran, H.
\newblock (2018).
\newblock Autonomous vehicle perception: The technology of today and tomorrow.
\newblock {\em Transportation research part C: emerging technologies},
  89:384--406.

\bibitem[\protect\citename{Wang \bgroup et al.\egroup }2013]{Wang_CER}
Wang, P., Sun, R., Zhao, H., and Yu, K.
\newblock (2013).
\newblock A new word language model evaluation metric for character based
  languages.
\newblock In Maosong Sun, et~al., editors, {\em Chinese Computational
  Linguistics and Natural Language Processing Based on Naturally Annotated Big
  Data}, pages 315--324, Berlin, Heidelberg. Springer Berlin Heidelberg.

\bibitem[\protect\citename{Wang \bgroup et al.\egroup }2020]{259725}
Wang, S., Cao, J., Sun, K., and Li, Q.
\newblock (2020).
\newblock {SIEVE}: Secure {In-Vehicle} automatic speech recognition systems.
\newblock In {\em 23rd International Symposium on Research in Attacks,
  Intrusions and Defenses (RAID 2020)}, pages 365--379, San Sebastian, October.
  USENIX Association.

\bibitem[\protect\citename{Winata \bgroup et al.\egroup }2020a]{winata2020lrt}
Winata, G.~I., Cahyawijaya, S., Lin, Z., Liu, Z., and Fung, P.
\newblock (2020a).
\newblock Lightweight and efficient end-to-end speech recognition using
  low-rank transformer.
\newblock {\em ICASSP 2020 - 2020 IEEE International Conference on Acoustics,
  Speech and Signal Processing (ICASSP)}, pages 6144--6148.

\bibitem[\protect\citename{Winata \bgroup et al.\egroup }2020b]{winata2020mtl}
Winata, G.~I., Cahyawijaya, S., Lin, Z., Liu, Z., Xu, P., and Fung, P.
\newblock (2020b).
\newblock Meta-transfer learning for code-switched speech recognition.
\newblock In {\em Proceedings of the 58th Annual Meeting of the Association for
  Computational Linguistics}, pages 3770--3776, Online, July. Association for
  Computational Linguistics.

\bibitem[\protect\citename{Xia \bgroup et al.\egroup
  }2020]{Xia2020AudiovisualSR}
Xia, L., Chen, G., Xu, X., Cui, J., and Gao, Y.
\newblock (2020).
\newblock Audiovisual speech recognition: A review and forecast.
\newblock {\em International Journal of Advanced Robotic Systems}, 17.

\bibitem[\protect\citename{Xu \bgroup et al.\egroup
  }2020]{xu2020discriminative}
Xu, B., Lu, C., Guo, Y., and Wang, J.
\newblock (2020).
\newblock Discriminative multi-modality speech recognition.
\newblock In {\em Proceedings of the IEEE/CVF Conference on Computer Vision and
  Pattern Recognition}, pages 14433--14442.

\bibitem[\protect\citename{Zadeh \bgroup et al.\egroup }2019]{zadeh2019fmt}
Zadeh, A., Mao, C., Shi, K., Zhang, Y., Liang, P.~P., Poria, S., and Morency,
  L.
\newblock (2019).
\newblock Factorized multimodal transformer for multimodal sequential learning.
\newblock {\em CoRR}, abs/1911.09826.

\bibitem[\protect\citename{Zhou \bgroup et al.\egroup }2019]{zhou2019hidden}
Zhou, M., Qin, Z., Lin, X., Hu, S., Wang, Q., and Ren, K.
\newblock (2019).
\newblock Hidden voice commands: Attacks and defenses on the vcs of autonomous
  driving cars.
\newblock {\em IEEE Wireless Communications}, 26(5):128--133.

\bibitem[\protect\citename{Zhu \bgroup et al.\egroup }2021]{zhu2021deep}
Zhu, H., Luo, M.-D., Wang, R., Zheng, A.-H., and He, R.
\newblock (2021).
\newblock Deep audio-visual learning: A survey.
\newblock {\em International Journal of Automation and Computing}, pages 1--26.

\end{thebibliography}

\end{document}